# Advancing a taxonomy for proxemics in robot social navigation

Ehud Nahum*, Yael Edan and Tal Oron-Gilad

Dept. of Industrial Engineering and Management
Agricultural, Biological, Cognitive Robotics Initiative
Ben-Gurion University of the Negev, Beer-Sheva, Israel

*Corresponding author: ehudna@bgu.ac.il.
Contributing authors: yael@bgu.ac.il; orontal@bgu.ac.il.
These authors contributed equally to this work.

## Abstract

Deploying robots in human environments requires effective social robot navigation. This article focuses on proxemics, proposing a new taxonomy and suggesting future directions through an analysis of state-of-the-art studies and the identification of research gaps. The various factors that affect the dynamic properties of proxemics patterns in human-robot interaction are thoroughly explored. To establish a coherent proxemics framework, we identified and organized the key parameters and attributes that shape proxemics behavior. Building on this framework, we introduce a novel approach to define proxemics in robot navigation, emphasizing the significant attributes that influence its structure and size. This leads to the development of a new taxonomy that serves as a foundation for guiding future research and development. Our findings underscore the complexity of defining personal distance, revealing it as a complex, multi-dimensional challenge. Furthermore, we highlight the flexible and dynamic nature of personal zone boundaries, which should be adaptable to different contexts and circumstances. Additionally, we propose a new layer for implementing proxemics in the navigation of social robots.

**Keywords:** Proxemics, Human-aware navigation, Socially-aware navigation, Social-robot navigation, Taxonomy, Human-robot interaction.

## 1 Introduction

Social robots must navigate with and among people in a safe and socially acceptable manner, particularly in shared spaces [1]. "A socially aware navigation is the strategy exhibited by a social robot which identifies and follows social conventions (regarding physical space management) to preserve a comfortable interaction with humans." (page 146) [2].

A socially navigating robot is defined as one that "acts and interacts with humans or other robots, achieving its navigation goals while modifying its behavior so the experience of agents around the robot is not degraded or is even enhanced" [1] (page 3). To navigate effectively in diverse physical and social environments, robots must be equipped with mechanisms that account for human behavior patterns and preferences in shared activities [3]. Specifically, socially aware robot navigation requires robots to understand how humans perceive and manage shared spaces. A key component of this process is enabling robots to interpret and adhere to social rules, such as proxemics - a foundational framework for studying human spatial behavior – guiding their navigation and interactions with humans in physical spaces.

Proxemics, originally developed in psychology, explores how individuals perceive and manage physical space in relation to others. This concept has since evolved to encompass the design of robot behavior in social contexts. The physical proximity of robots to humans, and the space humans perceive around them highlights the importance of adapting robots' navigation to align with human proxemics and social norms. While researchers widely agree that robots should respect social norms regarding physical and psychological distance, this understanding has yet to be fully implemented in practice.

Some studies define "personal distance" as the minimum spatial separation required between humans and robots, considering factors like time and space. However, despite these efforts, the practical implementation of these insights is still in the early stages.

People perceive and organize the space and distance they maintain between one another based on culturally specific situations [4] with personal space being a key concept in proxemics. Humans subconsciously consider other humans' personal space when interacting socially and navigating an environment [5]. The idea of proxemics between humans was introduced in 1966 by Hall [2] who proposed a fixed measure for personal space, defining it as a set of circular regions around a person. In this model, the expected distance and intimacy define four regions (Fig. 1) for interaction and communication with other people. This proxemics theory has been used in multiple research studies [6], [7], [8], [9], [10], [11], [12] with proxemics shape defined in different ways. Some researchers have based their work on Hall's circular model, including Repiso et al.[13], Bilen et al.[14], Karwowski et al.[15], Singh et al.[7], and Hanumantha et al.[8]. Other researchers have proposed new forms influenced by human behaviors and other factors Kang et al., Clavero et al. [16], [17].

In this paper, we extend this theory and present the need for a more complex approach for defining proxemics in the social navigation of robots.

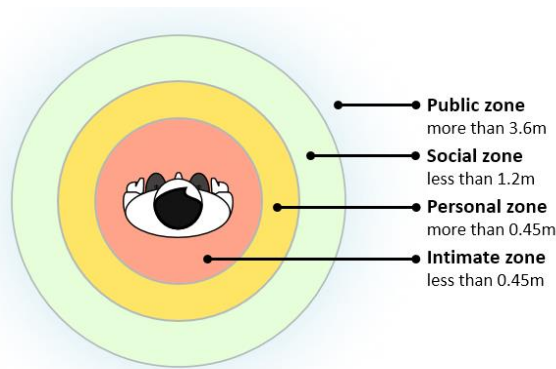

**Fig. 1** Hall's circular proxemics zones [2]

For humans and robots to share space and interact effectively, robot navigation must consider social factors and human comfort (or discomfort) [18] while avoiding inappropriate interaction conditions. Researchers widely agree that robots should adhere to social norms [6], [14], [19], [20], [21] concerning physical and psychological distance to ensure successful interaction with humans. However, transforming this consensus into practical applications remains an open challenge. Some studies define "personal distance" as the minimum spatial separation required between humans and robots, considering factors such as time, space, and context. However, proxemics-related behavior in robots is still in its early stages of implementation.

Studying proxemics in HRI is necessary to develop socially aware robot navigation [15]. Many studies and experiments describe the proxemics properties as a static, fixed area around a human affected by various factors [6], [8], [10], [12], [16], [20], [22], [23], [24], [25], [26]. One of the first attempts to define a proxemic as a dynamic zone that changes shape and size during the interaction depending on human behavior and context [15] defined four distinguishable proxemic shapes (Fig. 2).

The "Centric cycles" with the human at the center is the classic shape aligned with Hall's theory [2] (as shown in Fig. 1). Its symmetric shape is simple to implement, and the authors noted that its circumference is affected by factors including age, culture, type of relationship between interacting humans or agents (human-robot), and the context.

The 'Egg shape' dictates different front and back distances from the human with a more considerable distance upfront[27].

In the ellipse shape, the human is in the middle of the shape with an equal distance from the front and back and a second radius for the left and right sides of the human. This "Concentric ellipse," proposed by Helbing [28] is based on dynamics and pedestrian movements.

The fourth shape is the "Asymmetric shape" [29], which refers to a non-symmetrical zone where the human's dominant hand (e.g., right side) requires less space than the left side. It corresponds to [29] where the distance is smaller on the pedestrian-dominant side and does not vary according to the person's walking speed, these four shapes illustrate the complexity of defining proxemics shapes. However, they still do not account for dynamic properties, appropriateness, or factors like speed, appearance, or approach direction.

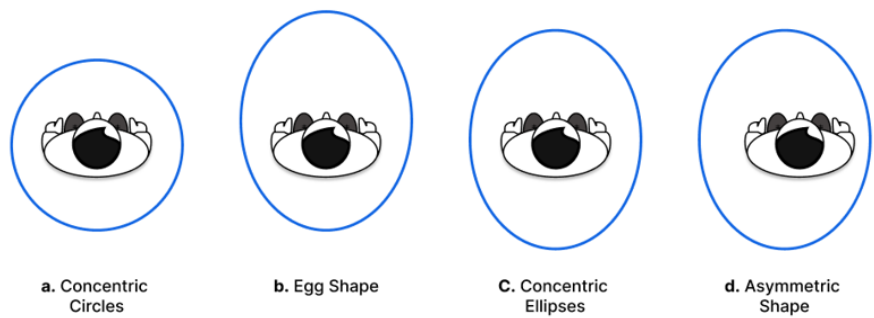

**Fig. 2** Proxemics shapes in HRI based on survey [15]: Concentric Circles [2], Egg Shape [27], Concentric Ellipses [28] and Asymmetric shape [29].

The most significant impact of proxemics occurs within the personal zone, where the physical domination is restricted. People can interact naturally in this zone, maintaining a distance that barely allows close contact. Beyond this zone, it becomes more difficult for a person to engage with somebody else [24]. Our work, therefore, aims to focus on what Hall defined as the personal zone, delimited by distances of 0.45m to 1.2 m. [2]. Finally, we propose a new method of attribute classification that emphasizes how each attribute influences the proxemic zone.

In the current study, we map and highlight the influence of various parameters and attributes on the shaping of proxemics in human-robot interaction (HRI), as well as their dynamic properties, based on a comprehensive literature review. The review encompasses a five-year analysis of studies on robot navigation among people, spanning from January 2020 to December 2024. The selected articles include both experimental work and simulated use cases in socially aware robot navigation that focus on the robot's proxemics-related behavior. The objectives of this review were to examine the experimental methodologies used across these studies, understand how proxemics shape is defined, identify the key attributes that influence its formation, and characterize proxemics through both quantitative and qualitative analyses. Based on this review, we propose a novel approach to defining proxemics in robot navigation and highlight the key attributes shape and determine its dimensions. This led to the development of a new taxonomy, which serves as a framework for mapping the various studies and identifying research gaps. Additionally, it provides guidance for future research and development in the field.

The work is organized as follows. Section 2 presents our methodology and a review of the work done in the last five years. In Section 3, we define the concept of the proxemics shape, followed by a proposal for a new proxemics taxonomy in Section 4. Section 5 provides an overview of the relevant literature in this context. Finally, we conclude the review by discussing the limitations of existing proxemics research, highlighting key challenges, and identifying opportunities for future exploration.

## 2 Literature review methodology

The following review included only human-robot interaction studies that addressed proxemics between January 2020 until December 2024. The survey specially examines on HRI studies that incorporated experimental data for ground robots, mobile wheeled and legged, and humanoid robots [30] across various environments- physical, simulation, and virtual.

The following three stages were applied for scoping and conducting this literature review (Fig. 3): Identification, screening, eligibility, and inclusion.

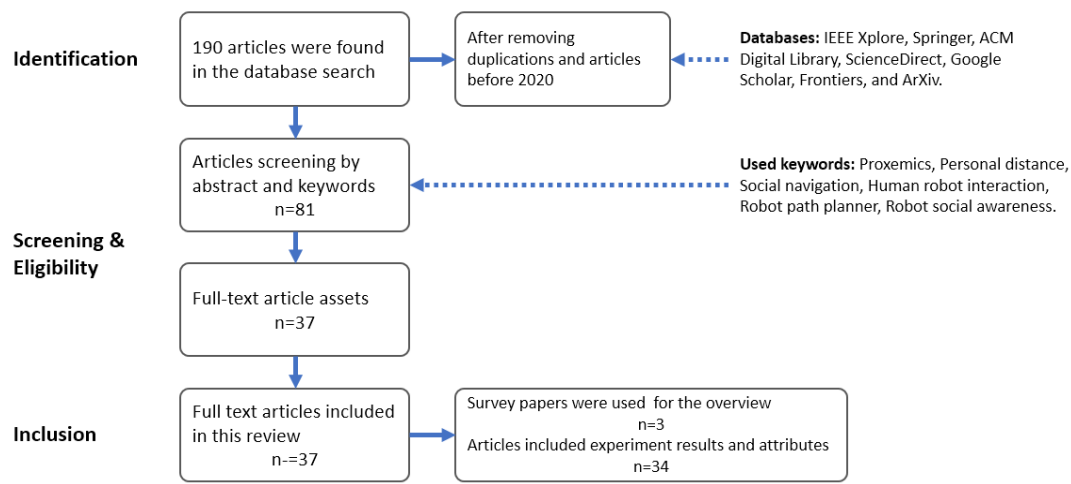

**Fig. 3** Diagram of article search and screening process for inclusion in the review.

**Identification:** for the literature search, we used the following query keywords: Proxemics, Personal distance, Human-robot interaction, Social aware navigation, and Robot path planner. We defined the boundaries of this study from the keywords, refined the preliminary keywords and literature screening, and excluded irrelevant articles.

We sourced papers and articles on human-robot proxemics through the following databases: IEEE Xplore, Springer, ACM Digital Library, ScienceDirect, Google Scholar, Frontiers, and ArXiv. The literature library compiled as part of this review contains science publications, magazine articles, conference publications, and thesis works. This process yielded more than 190 articles in the initial search.

**Screening and eligibility:** From the above, we isolated 81 articles published after January 2020 (Fig. 4). Then, we excluded all theoretical and opinion papers, resulting in 37 articles (Fig. 3) that included empirical studies or survey papers.

**Inclusion:** The 37 included papers are organized into two groups (Fig. 3): Survey papers (3 papers) used to derive key factors of proxemics. Experimental articles (a total of 34 publications) were tagged manually and classified into four classifications reflecting the taxonomy's main categories: robot, human, environment, and context.

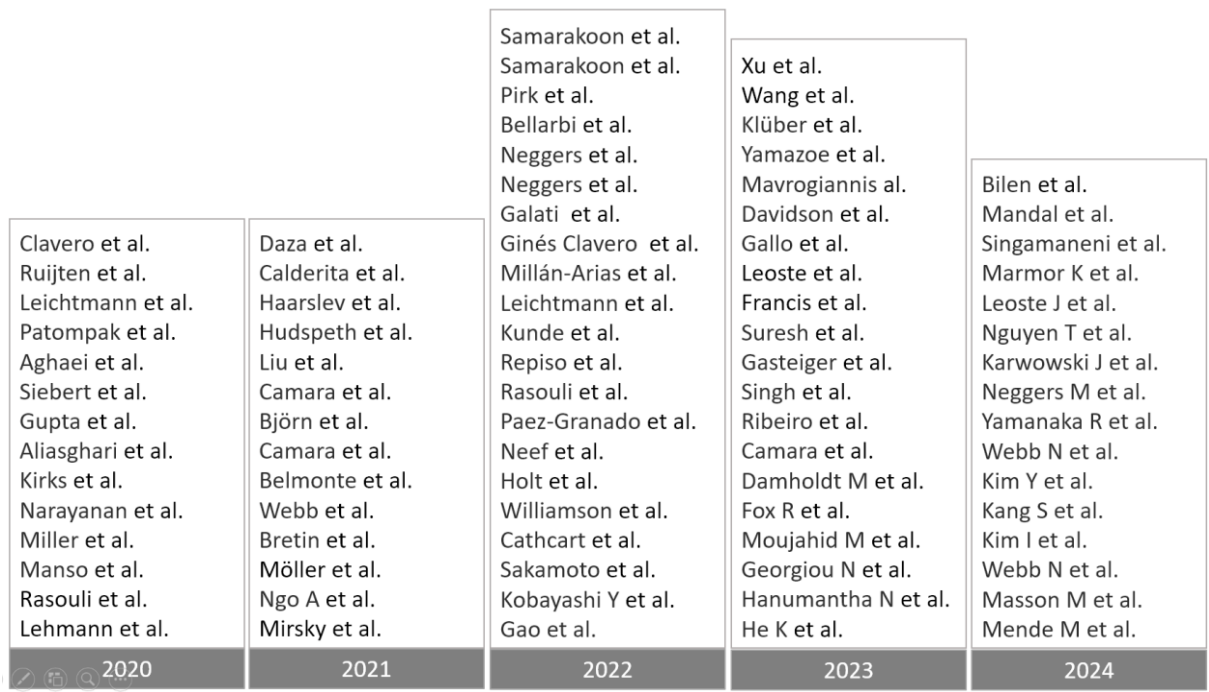

**Fig. 4** Distribution by years (Jan 2020 to Dec 2024) of social navigation papers focusing on proxemics.

# 3 Review of proxemics in robot social navigation

## 3.1 Previous surveys

We begin with selected reviews by Samarakoon et al. (2022) [31], Mavrogiannis et al. (2023), and Singamaneni et al. (2024) [30] and the article by Clavero et al. (2020) [17]. The reviews and the article were used to identify distinctive directions that would form a representative subset of what has been done in proxemics. We first describe the state-of-the-art proxemics research in social robot navigation derived from the reviews.

Samarakoon et al. (2022)[31]focused on taxonomizing and analyzing both human-human proxemics and human-robot proxemics and proposed a method for establishing human-robot proxemics. Their framework for human-human proxemics builds on several key studies: Hall's theory with the classic four zones (Fig. 1), Argyle and Dean's experiment [32], which explored the relationship between eye contact and interpersonal distance, revealing that eye contact influences the proxemics distance, and Kendon and Ciolek work [33] on six basic F-formations that offer a structured arrangement of people for specific tasks. The research highlights various situations based on these six F-formations for pairs of individuals. For human-robot proxemics, the authors incorporated user attributes from three studies: Walters et al. [34], who measured preferred distance for adults and children as they approached a robot; Obaid et al. [35], who found no significant effect of gender; and Eresha et al. [22], who examined the impact of user culture[36].

Mavrogiannis et al. (2023) [37] provide a comprehensive view of proxemics spanning from Hall's traditional theory to the Hayduk and Kirby egg shape space [27], [38]. Gérin-Lajoie views the shape as asymmetric [29], and Hayduk considers it dynamic. The authors argue that a major proxemic barrier lies in the limited understanding of human pedestrian behavior. This gap in knowledge raises several questions: How should personal space be modelled, and how can robots represent this space? How should the robot communicate its intent? and how can a robot identify and interpret a group of pedestrians in context?

In Singamaneni et al. (2024)[30], proximity is part of a situation awareness taxonomy. The authors note that while current analyses and recent surveys of socially aware robot navigation cover many key aspects of the field, they overlook many essential elements, such as the different types of robots and how their specific characteristics might influence navigation strategies. Additionally, the effects of social context and the semantics of the environment are often missing. Their proposed taxonomy aims to fill these gaps by including all relevant factors that could be present in existing or future approaches to socially aware robot navigation. The authors also identify future challenges, emphasizing that designing **user-adaptive robot behavior** requires improving the robot's perception of humans. This should include robot-specific parameters to adapt to different robot characteristics, as well as exploring better alternatives for proxemics-based metrics and the need to consider additional factors that affect user comfort around robots.

Further, the literature by Clavero et al. (2020) [17] presents two main contributions to the above reviews. Their theoretical contribution defines proxemics as an adaptive shape influenced by human activity, location, and culture. This new proxemics shape represents a **cooperation zone** that allows natural cooperation between humans and robots during navigation and interaction tasks. Their technical contribution involves a ROS2 navigation layer that maps a dynamic proxemic zone on a map, enabling developers to create complex behaviors. Specifically, the authors propose an asymmetric Gaussian proxemics zone defined by four variables: front, back, and sides (which use the same value) to create an adaptive zone with varying shapes and sizes. They argue that the proxemics zone should be defined according to the human's activity (e.g., a different shape for running, or standing), and the shape should be represented as an Asymmetric Gaussian function.

### 3.2 Proxemics shape

Proxemics between humans was introduced by Hall in 1966 [2] who proposed a fixed measure for personal space as a set of circular regions around a person. The expected distance and intimacy define four regions (Fig. 1) during interaction and communication with other people.

Some current works agree with Hall's method [6], [7], [12], [13], claiming that the personal space is circular Hecht et al. [39] and with the categorization of zones as related to different types of interactions (Fig. 1). Other works found that personal space is a non-circular shape [40] (e.g., Fig. 2). Some people, as was tested in real-world environments, need more space on the frontal side [27]. The explanation for requiring more space in front is the variety of actions a human can perform, such as standing and walking, or activities within the social moderators, like eye gaze. Eye contact in interpersonal distance prevents the interaction from becoming intimate, and therefore, when someone comes from the back, the personal distance can be smaller [41]. On the contrary, [42] found that people need more space behind them. In this study, when another human approached the participant from the back, they needed more space than when approached from other sides. This can be explained by having a lean sensory cover on the back while having a vision sensory to control the required distance in the front.

Most models to date defined proxemics while examining the approach to a static person (20). However, one must also consider the direction and velocity when a human is in movement. Helbing et al. [28] modeled three social forces that influence the motion direction of pedestrians, which result in an elliptical shape. These forces are Desired Velocity, where pedestrians maintain a preferred speed and direction they want to move toward; Repulsive Forces, where pedestrians maintain a certain distance from others and obstacles to avoid collisions; and Attractive Forces, where pedestrians are drawn toward certain goals or destinations.

Some works (4 of the 81 papers) noted an asymmetric personal space shape. In the work of Gerin-Lajoie et al. [29] where participants were asked to bypass the obstacle, participants kept a smaller distance on their dominant hand side. This behavior is also observed in corridors as pedestrians usually pass other people on the right side of the corridor. These results indicate that it is largely unclear which shape best represents the human comfort zone.

### 3.3 Proxemics evaluation

Most of the studies included in the review are lab studies (25 papers) that are heavily controlled and lack authenticity compared to real-world scenarios. A real-world outdoor study was performed only in a single paper [13].

Some of the works are based on human-human interaction as in Liu et al. [5], some are tested in virtual [43], or simulated environments [6], [8], [10], [16], [20], [21], [22], [24], [25] some in real indoor environments [5], [7], [9], [10], [11], [12], [16], [19], [20], [22], [23], [25], [44], [45], [46], [47], [48], [49], [50], [51], [52], [53], [54], [55], [56], [57], some check humans in static positions [5], [7], [9], [10], [11], [12], [13], [16], [19], [22], [23], [24], [25], [43], [46], [47], [48], [52], [54], [56], [58], and some while walking [6], [7], [9], [13], [44], [45], [48], [50], [51], [53], [55], [58].

In 23 of the studies, only one attribute is tested. Six papers dealt with two parameters [5], [44], [45], [49], [50], [55], and five papers analyzed more than two parameters [10], [13], [23], [46], [54].

However, as shown in Fig. 6, since the shape of the personal space depends on various variables such as task content, speed and direction of movement, personal characteristics, and social cues, these must be analyzed in user studies.

Most of the metrics presented focus on the efficiency and safety of the robot: distance (23), directions (11), speed (6), path length (7), safety (5), navigation (4), success rate (1), path efficiency (1), collision detection (2), and behavior naturalness (1). In contrast, social-related metrics, which include gaze (3), comfort (11), trust (3), emotional experience (3), and satisfaction (1) imply safety or overall ratings of human participants' perceptions.

## 4 A proposed taxonomy for proxemics in HRI

### 4.1 Method

The methodological flow to derive the proposed taxonomy is described in Fig. 5 and detailed below.

Our proposed taxonomy is derived from a mapping of the attributes that influence the shape and size of proxemics, based on insights from the literature review. We then mapped the 34 selected articles to the relevant components of the taxonomy, aiming to highlighting well-explored topics and identify gaps.

To begin the classification, we used the HRI taxonomy by Onnasch et al. (2020) [59] (see Table 1) In this article, the taxonomy is organized into three clusters: Context, Robot, and Team. The author describes this taxonomy as providing "a structural basis to classify various aspects of HRI". We evaluated the articles classification, and we found that the suggested classification uncovers the proxemics related attributes.

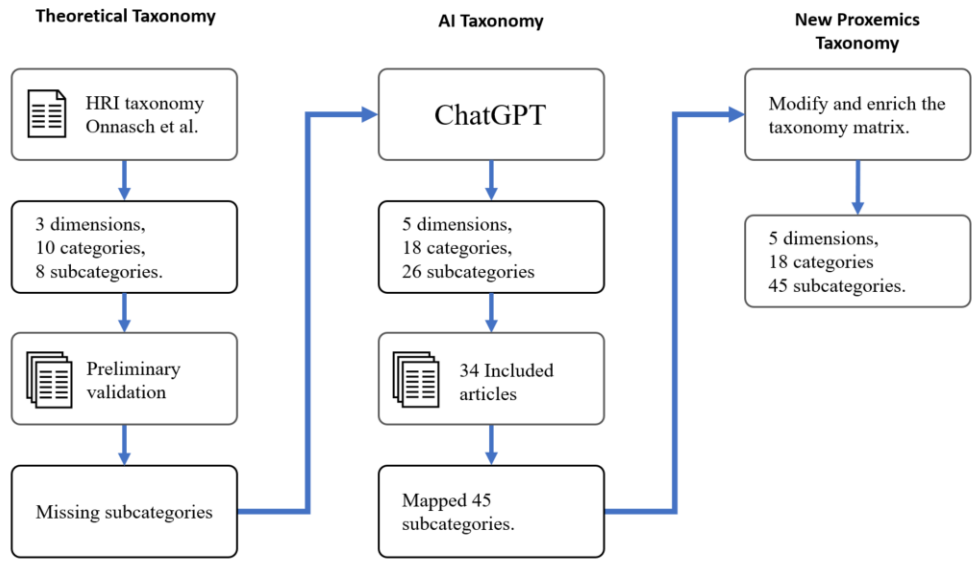

**Fig. 5** Taxonomy classification flow – Theoretical to practical.

In order to create a comprehensive overview of related taxonomy entities, we sought to develop an innovative and inspired classification as a starting point. We explored AI tools like Gemini and Copilot, ultimately selecting ChatGPT to conduct an extensive mapping of the taxonomy for HRI social navigation. In January 2025, we entered the following queries: "Can you provide a structured table of HRI proxemics taxonomy, organized hierarchically based on Human, Robot, Environment, and Context?" then we asked, "Can you provide it to HRI proxemics?". The resulting output identified five dimensions with detailed categories and subcategories (see Table 2). We then integrated this classification into our taxonomy mapping.

**Table 1** Onnasch et al. [59] taxonomy used as preliminary HRI classification.

| Dimensions | Category | Subcategory | Details |
|---|---|---|---|
| Interaction Context Classification | Field of Application | - | Industry, Service, Military & Police, Space Expedition, Therapy, Education, Entertainment, None |
| | Exposure to the Robot | - | Embodied, Depicted |
| | Setting | - | Field, Laboratory |
| Robot Classification | Robot Task Specification | - | Information Exchange, Precision, Physical Load Reduction, Transport, Manipulation, Cognitive Stimulation, Emotional Stimulation, Physical Stimulation |
| | Robot Morphology | Appearance | Anthropomorphic, Zoomorphic, Technical |
| | | Communication | Anthropomorphic, Zoomorphic, Technical |
| | | Movement | Anthropomorphic, Zoomorphic, Technical |
| | | Context | Anthropomorphic, Zoomorphic, Technical |
| | Degree of Robot Autonomy | - | Information Acquisition, Information Analysis, Decision-Making, Action Implementation |
| Team Classification | Human Role | - | Supervisor, Operator, Collaborator, Cooperator, Bystander |
| | Team Composition | - | Equal humans and robots, More humans than robots, More robots than humans |
| | | Input | Electronic, Mechanical, Acoustic, Optic |

| | Communication Channel | Output | Tactile, Acoustic, Visual |
|---|---|---|---|
| | Proximity | Physical | Following, Touching, Approaching, Passing, Avoidance, None |
| | | Temporal | Synchronous, Asynchronous |

Table 2 The ChatGPT results for the query to create an HRI social navigation taxonomy.

| Dimensions | Category | Subcategory | Details |
|---|---|---|---|
| Human | Demographics | - | Age, cultural background, physical abilities (e.g., children, elderly, disabled). |
| | Behavioral Patterns | - | Walking speed, group dynamics (e.g., solo or group navigation), adherence to social norms. |
| | Comfort and Trust | - | Perception of safety, comfort zones, and trust in robot capabilities. |
| | Communication Preferences | Verbal | Speech-based communication. |
| | | Non-verbal | Gestures or facial expressions. |
| | | Multimodal | Combination of verbal and non-verbal communication. |
| Robot | Capabilities | Sensing | LiDAR, cameras, motion prediction. |
| | | Path planning and actuation | Smooth movement. |
| | Behavior | Social norm adherence | Yielding, signaling, and avoiding collisions. |
| | | Signaling intent | Using LEDs, gestures, or sounds. |
| | Design | Physical size | Size, form factor. |
| | | Locomotion | Wheeled, legged, or aerial robots. |
| | Role | Guide | Leading individuals or groups. |
| | | Assistant | Supporting tasks. |
| | | Companion | Offering emotional or social interaction. |
| | | Delivery | Transporting goods. |
| Environment | Type of Space | Indoor | Homes, malls, offices. |
| | | Outdoor | Parks, streets. |
| | | Semi-structured | Airports, arenas. |
| | Crowd Dynamics | - | Low, medium, or high density and flow patterns. |
| | Obstacles | Static | Furniture or fixed objects. |
| | | Dynamic | Humans, carts, animals. |
| | Safety Requirements | - | Emergency protocols and accessibility features (e.g., ramps, wide aisles). |
| Context | Task Objective | - | Guidance, delivery, companionship, or monitoring. |
| | Time Sensitivity | Urgent | Emergency response tasks. |
| | | Non-urgent | Routine tasks like patrolling. |
| | Social Norm Adherence | - | Robot behavior aligns with cultural and local norms. |
| | Interaction Level | Interaction-focused | Engaging individuals. |
| | | Navigation-focused | Efficient movement in crowded spaces. |
| Proximity | Zones | Intimate | <0.5 m (e.g., close personal assistance). |
| | | Personal | 0.5–1.5 m (e.g., one-on-one interactions). |
| | | Social | 1.5–3 m (e.g., group navigation). |
| | | Public | >3 m (e.g., audience interaction). |
| | Dynamic Adjustments | - | Real-time changes in behavior based on movement, crowd density, and feedback. |

Our clustering method for the proxemics taxonomy followed a top-bottom approach. We extracted the attributes from the 34 reviewed papers and organized them into four clusters: Human, Robot, Environment, and Context (Fig. 6), based on the ChatGPT mapping taxonomy (Table 2). This extraction process uncovered attributes that were not included in the existing mapping (for example, human skill and experience).

We then used a bottom-up approach to refine the mapping to improve our taxonomy definition. The result is our proposed taxonomy, which is presented in Fig. 7-11. Recall that in most of the papers (23), proxemics experiments focused on a single attribute, such as robot height (1), human position (25), and robot design (33). In some cases, two attributes were analyzed, including approach directions (9), velocity (3), and the presence of individuals vs. groups (5). These collected attributes were used to define a set of taxa.

Fig. 6 presents the resulting taxonomy – a top-level description of the proposed clusters and their related taxa.

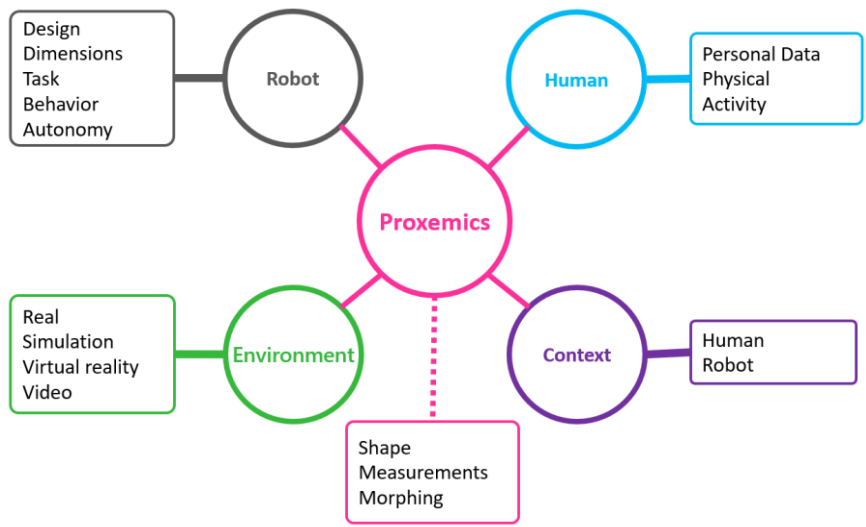

**Fig. 6** An overview of the proxemics taxonomy.

In the following, we detail each taxa derived from the literature review.

### 4.2 Taxonomy for Robot.

Various attributes are included when we deal with social robots. As noted, we focus only on articles related to ground robots, robots that move and place on the ground.

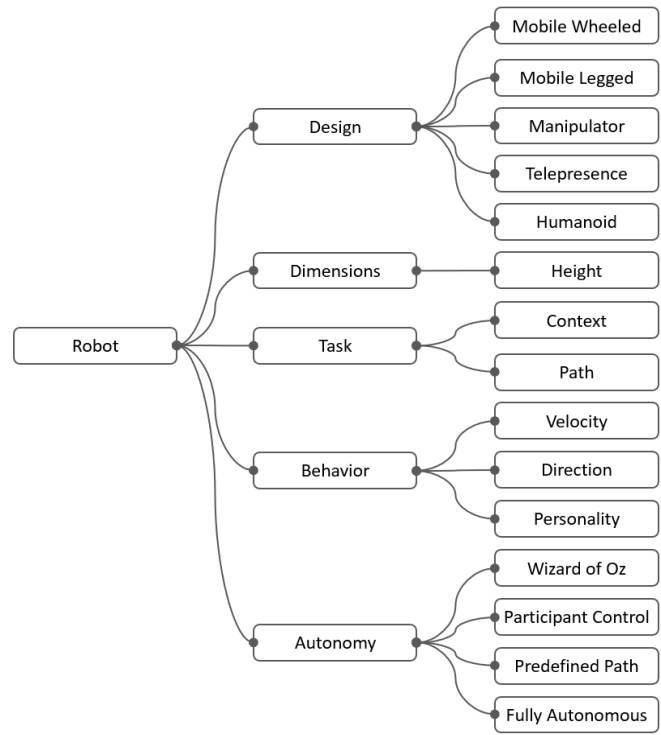

**Fig. 7** Taxonomy for robot characteristics, tasks, behavior, and control.

### 4.2.1 Definitions
1) **Design***:* This taxon relates to the robot's type and movability mechanism.
    a) **Mobile robot:** any robot that moves with wheels and looks like a machine.
        - **Wheels**: any robot that moves with wheels.
        - **Legs:** any robot that moves with legs.
    b) **Manipulator**: a mobile robot designed to perform physical manipulation tasks with its arm.
    c) **Telepresence:** any robot that enables users to maintain a virtual presence in a remote location. The user can control the robot's movements and other behavior through a computer.
    d) **Humanoid**: a robot shaped like a human (body and head, sometimes even hands).
2) **Dimensions** are the robot's physical measurements that affect its behavior.
    a) **Height:** the measurement from the floor to the highest point of the robot.
    b) **Footprint:** the space that the robot takes on the ground
    c) **Massive:** robot shape mass impression, whether heavy or fragile.
3) Task: this taxon classified the mission/work the robot ordered to do.
    a) **Context**: the type of
    b) **Path**: the trajectory from the origin to the target point.
4) Behavior:
    a) **Velocity**: the moving speed related to the task priority and the scene limitation.
    b) **Direction**: the side/angle when a robot approaches the person for interaction.
    c) **Personality**: A reflection of human emotional patterns such as happy, angry, etc.

5) **Autonomy**
   a) **Wizard of Oz:** when an operator controls the robot behind the scenes, and the participant isn't aware.
   b) **Participant control:** when the participant moves the robot with a remote control during the experiment.
   c) **Predefined Path:** when a robot moves between two predefined points, starting and destination.
   d) **Fully autonomous:** a robot that can sense and move among humans in the same environment.

### 4.2.2 Related work.

**Table 3** Attributing the review papers to the related robot taxa for design, dimensions, task, behavior and autonomy.

| Category | Attribute | Articles |
|---|---|---|
| Design | Mobile wheels | [6], [9], [10], [12], [20], [21], [22], [23], [24], [43], [46], [47], [51], [52], [53], [55], [58] |
| | Mobile legs | [45] |
| | Manipulator | [8], [16], [50], [56] |
| | Telepresence | [7], [19], [48], [52] |
| | Humanoid | [5], [11], [13], [23], [25], [44], [49], [51], [54], [57], [60], [61] |
| Dimensions | Height | [56] |
| Task | Context | [5], [7], [9], [19], [24], [48], [49], [56], [58] |
| | Path | [50] |
| Behavior | Velocity | [43], [44], [46], [51] |
| | Directions | [5], [6], [23], [44], [45], [46], [50], [51], [52], [54], [55] |
| | Personality | [49] |
| Autonomy | Wizard of Oz | [5], [19], [47], [52], [56] |
| | Participant control | [21], [48] |
| | Predefined path | [10], [20], [43], [44], [46], [53], [54] |
| | Fully autonomous | [7], [9], [13], [22], [23], [45], [51], [52], [55], [57] |

## 4.3 Taxonomy for human
This classification contains the personal, physical, and activity of the human.

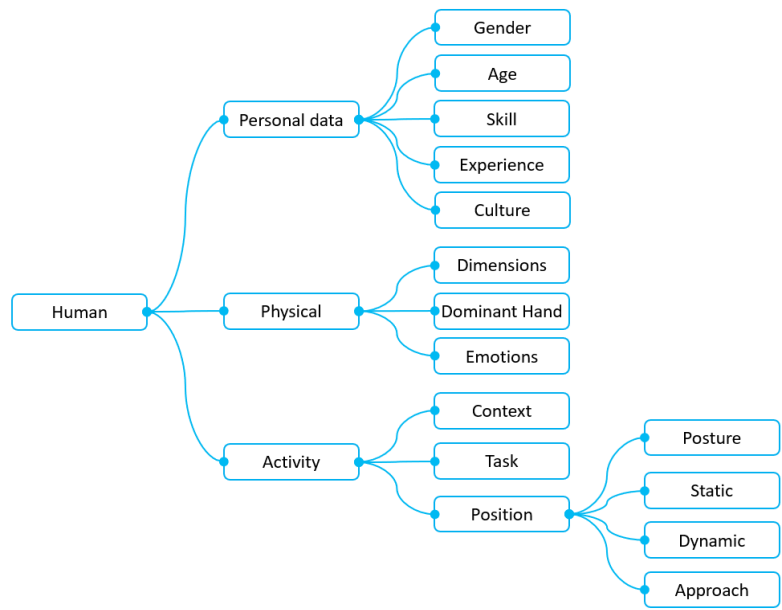

**Fig. 8** Taxonomy for human personal attributes and activity.

### 4.3.1 Definitions
1) **Personal data:** This taxon is related to human physical and personal characteristics and behavioral aspects.
    a) **Gender**: the human's sex type that will affect his
    b) **Age:** human age affects his physical dimensions and behaviors.
    c) **Skill:** if the human has a specific experience that differentiates him from others.
    d) **Experience:** If the participant has an early experience with robots.
    e) **Culture:** related regions and cultures have different social norms.
2) Physical
    a) **Dimension:** the human height that, in most cases, is derived from his age.
    b) **Dominant hand:** the stronger hand of the human.
    c) **Emotions**: the human mood affects his social behavior.
3) Activity
    a) **Context:** when an interaction has a scenario based on a real-world situation, such as the delivery of an item.
    b) **Task:** while the human has a specific activity or mission to do.
    c) **Position**
        - **Posture:** the body poses during the interaction, such as standing or sitting.
        - **Statics:** when the human is located at the same stationary point during the experiment.
        - **Dynamic:** when the human moves from the experiment starting point, such as walking or changing his position.
        - **Approach:** When a human comes close to interacting with the robot.

### 4.3.2 Related work.

**Table 4** Attributing the review papers to the related human taxa for personal data, physical properties and activity.

| Category | Attribute | Articles |
|---|---|---|
| Personal data | Gender | [43], [45], [48] |
| | Age | [20] |
| | Skill | [8], [43], [48] |
| | Experience | [56] |
| | Culture | |
| Physical | Dimensions | [20], [45] |
| | Dominant hand | |
| | Emotions | [61] |
| Activity | Context | [10], [13], [19], [47], [48], [49], [54] |
| | Task | [21], [45] |
| | Posture | [9], [43] |
| | Static | [5], [7], [9], [10], [11], [12], [13], [16], [19], [22], [23], [24], [25], [43], [46], [47], [48], [52], [54], [56], [58] |
| | Dynamic | [6], [7], [9], [13], [44], [45], [48], [50], [51], [53], [55], [58] |
| | Approach | [49], [57] |

## 4.4 Taxonomy for the environment

This classification considers the scene where the interaction happened and includes the scene type with its related attributes.

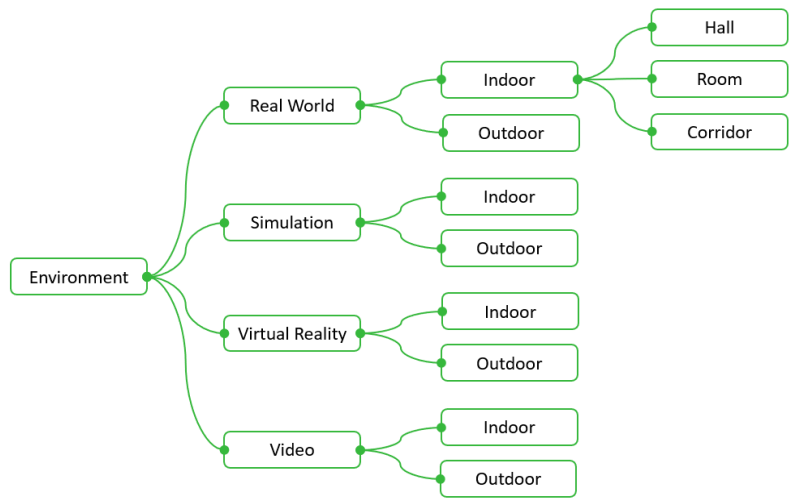

**Fig. 9** Taxonomy based on studies environments.

### 4.4.1 Definitions

1) **Real World:** This taxon contains articles that conduct experiments in a physical set with participants and robots.
    a) **Indoor**
        - **Hall:** An open space defined by its large size.
        - **Room:** a small space blocked by the walls, can be a lab or learning room.

- **Corridor:** an indoor lane used to move between places. It is blocked by a wall on its sides, with an average distance of 0.9 to 1.2 meters.
  b) **Outdoor**
    - **Sidewalk:** a public pathway shared with other people.
    - **Public space:** an open public space.
2) **Simulation:** This taxon grouped all papers that conducted studies and experiments in a computer-based simulation environment.
    a) **Indoor**
       - **Hall:** An open space defined by its large size.
       - **Room:** a small space blocked by the walls, e.g., a lab or learning room.
       - **Corridor:** an indoor lane used to move between places. It is blocked by a wall on its sides, with an average distance of 0.9 to 1.2 meters.
3) **Virtual Reality:** This taxon contains conducted experiments that allow the researcher to isolate the participant from environmental interference with complete scene control.
    a) **Indoor:** a closed space blocked between walls and ceiling.
    b) **Outdoor:** in most cases, it is an open public space.
4) **Video:** This taxon contains experiments conducted when the participant watches video clips and gives feedback later.
    a) **Indoor:** a closed space blocked between walls and ceiling.
    b) **Outdoor:** in most cases, it is an open public space.

### 4.4.2 Related work

**Table 5** Attributing the review papers to the related robot taxa for environment type.

| Category | Attribute | Articles |
|---|---|---|
| Real-world | Indoor | [5], [7], [9], [10], [11], [12], [16], [19], [20], [22], [23], [25], [44], [45], [46], [47], [48], [49], [50], [51], [52], [53], [54], [55], [56], [57] |
|  | Outdoor | [13] |
| Simulation | Indoor | [6], [8], [10], [16], [20], [21], [22], [24], [25] |
|  | Outdoor | [13] |
| Virtual reality | indoor | [43] |
|  | Outdoor |  |
| Video | Indoor | [11], [50], [58], [61] |
|  | Outdoor |  |

## 4.5 Taxonomy for the context

This classification contains experiments and papers in which the design demonstrates an actual situation.

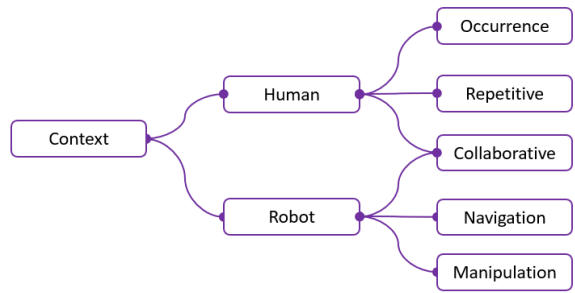

**Fig 10** Taxonomy for human's and robot's context.

### 4.5.1 Definition
1) **Human:** This taxon covered experiments with context or task while the participant did a specific activity.
    a) **Occurrence:** a single activity that happened once in the integration.
    b) **Repetitive:** a task that happens several times during the interaction.
    c) **Collaboration:** a task the human and robot are sharing.
2) **Robot:** a taxon while the robot was ordered to do a task.
    a) **Navigation:** when a robot moves through space from one waypoint to another.
    b) **Manipulation:** The robot has a mission to do some activity.

### 4.5.2 Related work
**Table 6** Attributing the review papers to the related robot taxa for context

| Category | Attribute | Articles |
|---|---|---|
| Human | Occurrence | [7], [9], [10], [21], [22], [25], [47], [50], [53], [54] |
| | Repetitive | [5], [56] |
| Both | Collaboration | [19], [49], [57], [58] |
| Robot | Navigation | [7], [8], [9], [10], [11], [13], [16], [20], [21], [23], [43], [46], [47], [50], [51], [52], [53], [54], [56] |
| | Manipulation | [22], [55] |

## 4.6 Taxonomy for the proxemics shape
This taxon includes experiments that contain proxemics measurements, some for a single direction, others for multiple directions, and some for defining the proxemics zone boundaries.

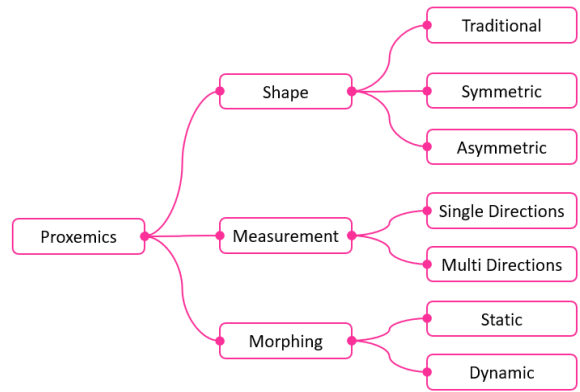

**Fig. 11** Taxonomy for proxemics shape, size, and morphing.

### 4.6.1 Definition
1) **Shape:** This taxon defines the geometry of the proxemics space footprint on the ground.
    a) **Traditional:** A circular shape based on Hall classification.
    b) **Symmetric:** A circular shape based on Hall's theory, the person is in the center, while the shape boundaries are equal distances from the person.
    c) **Asymmetric:** an oval shape while the participant is in the middle, and the distance from him to the shape boundaries is unequal.
    d) **Orientation:** the shape layout related to the participant's heading.
2) **Measurements:** This taxon describes the quantification of attributes that define the proxemics shape.
    a) **Single:** A one-directional measurement taken to define the shape, usually the forward distance.
    b) **Multiple:** More than one direction is measured to define the proxemics shape.
3) Morphing:
    a) **Static:** a predefined fixed shape size that is not changed during the interaction.
    b) **Dynamic:** a shape that changes size during interaction and the environment.

### 4.6.2 Related work
**Table 7** Attributing the review papers to the related robot taxa for the proxemics zone.

| Category | Attribute | Articles |
| --- | --- | --- |
| Shape | Traditional | [6], [7], [8], [9], [10], [11], [12] |
| | Symmetric | [24], [26] |
| | Asymmetric | [16], [20], [23], [25] |
| Measurements | Single direction | [9], [19], [21], [43], [48], [49], [56], [57] |
| | Multi directions | [5], [20], [23], [45], [46], [51], [52], [53], [54], [55] |
| Morphing | Static | [6], [8], [10], [12], [16], [20], [22], [23], [24], [25], [26] |
| | Dynamic | [9] |

# 5. Directions for future research.

Based on the analysis of the literature survey, we identify several missing attributes that are essential for defining a valid proxemics shape. Advancing experiments to explore these attributes is crucial. We summarize these key points as follows:

a) While the reviewed experiments and studies extensively covered robot design taxa, there is a noticeable gap in the exploration of robot dimensions and behaviors, which may significantly influence the proxemics zone.

b) Participant diversity is limited, with most studies primarily involving young people or students. A broader range of participants, incorporating factors such as age, culture, skills, and experience should be explored.

c) Many studies have been conducted in controlled, uniform environments, such as indoor rooms or lab corridors, with only one study examining an outdoor environment. It is important to advance experiments in a wide range of real-world environments.

d) Few studies account for context in their experiments. In most studies, robots simply move from point to point without a specific purpose. We argue that context is a critical factor that affects all aspects of HRI, particularly proxemics. Therefore, we recommend that future studies integrate contextual activities.

Social navigation is typically defined by three layers [62] that describe the interaction between a robot and a human in a shared environment [17]: the metric layer (defining spatial dimensions), the semantic layer (indicating where the robot can pass), and the social layer (indicating how the robot should move or overtake). These three layers are essential for the robot's perception and motion planning in dynamic environments. Researchers often use Hall's traditional proxemics zone to define the distance humans require when interacting with a robot. However, proxemics does not dominant this social layer.

To enhance proxemics, we propose an improvement to the existing social navigation layers. Specifically, we suggest adding an additional proxemics layer that bridges the semantic layer, where the human is situated, and the social layer, where the robot's movements occur (see (Figure 12). While the semantic and social layers primarily represent the robot's needs, the human aspects within these layers are designed to fulfill these needs. The proposed proxemics layer aims to integrate human-centered factors, incorporating elements related to human behavior and how it influences these layers. This vision behind this proxemics layer is to account for all human factors that affect and are affected by the interaction.

This new proxemics layer will combine the humans positions from the semantic layer with the human movement and behavior from the social layer. By integrating these parameters with the dynamic proxemics of the human, we can create a dynamic, integrated layer that reflects both human needs and behavior. This integration will enable the autonomous robot to better understanding human social norms while sharing the same space.

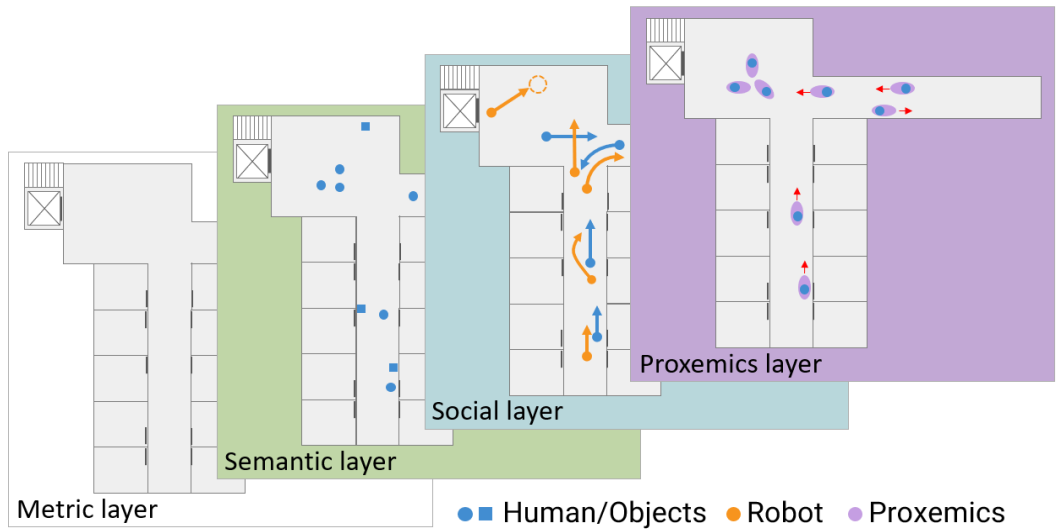

**Fig. 12** Social Navigation Mapping Layers based on Clavero et al. (2020) [17].

## 6. Conclusions.

Social navigation and proxemics have emerged as critical issues with the growing use of autonomous service robots in shared environments. For robots to be accepted by humans in shared spaces, they must exhibit sociable behaviors. The robot's actions directly influence the human sense of safety and trust.

We provide a new taxonomy for proxemics, offering a unique mapping and clustering approach. Unlike most previous surveys and studies in the field, which treat HRI social navigation as a collection of isolated issues, this paper defines the attributes influencing proxemics as asymmetric and dynamic, evolving throughout interactions with both humans and the environment. Our proposed taxonomy for proxemics in HRI accounts for all attributes that impact the shape and size of proxemics, providing a more systematic framework. Additionally, we introduce a new layer for defining proxemics within the context of social robot navigation.

We present a fresh perspective for future research, proposing a redefined approach to the proxemic zone as a dynamic shape that adapts to the human-robot environment and context. A flexible, dynamic shape that adjusts its boundaries based on various attributes is crucial for a deeper understanding of proxemics. As the field of social-navigating robots continues to advance and social robots become more prevalent in the market, incorporating these considerations will be crucial for shaping future developments.

## Acknowledgments

This research was supported by the Israeli Innovation Authority as part of the HRI consortium and partially supported by Ben-Gurion University of the Negev through the Rabbi W. Gunther Plaut Chair in Manufacturing Engineering, the George Shrut Chair for Human Performance, and the Agricultural, Biological Cognitive Robotics Initiative (funded by the Marcus Endowment Fund and the Helmsley Charitable Trust).

# References

[1] Y. Gao and C. M. Huang, "Evaluation of Socially-Aware Robot Navigation," Jan. 12, 2022, *Frontiers Media S.A.* doi: 10.3389/frobt.2021.721317.

[2] Hall, E.T. The Hidden Dimension; Doubleday & Company Inc.: Garden City, NY, USA, 1966.

[3] B. Kathleen *et al.*, "Addressing joint action challenges in HRI: Insights from psychology and philosophy," Feb. 01, 2022, *Elsevier B.V.* doi: 10.1016/j.actpsy.2021.103476.

[4] K. Brown, *Encyclopedia of language & linguistics*. 2006. doi: 10.5860/choice.43-5614.

[5] L. Liu, Y. Liu, and X. Z. Gao, "Impacts of human robot proxemics on human concentration-training games with humanoid robots," *Healthcare (Switzerland)*, vol. 9, no. 7, Jul. 2021, doi: 10.3390/healthcare9070894.

[6] T. T. Nguyen and T. D. Ngo, "Spatiotemporal Motion Profiles for Cost-Based Optimal Approaching Pose Estimation," in *2024 IEEE/SICE International Symposium on System Integration, SII 2024*, Institute of Electrical and Electronics Engineers Inc., 2024, pp. 92–98. doi: 10.1109/SII58957.2024.10417396.

[7] K. J. Singh, D. S. Kapoor, M. Abouhawwash, J. F. Al-Amri, S. Mahajan, and A. K. Pandit, "Behavior of Delivery Robot in Human-Robot Collaborative Spaces During Navigation," *Intelligent Automation and Soft Computing*, vol. 35, no. 1, pp. 795–810, 2023, doi: 10.32604/iasc.2023.025177.

[8] N. Hanumantha, "Improving The Autonomous Navigation Of A Care Robot By Following Social Norms In A Care Environment," 2023.

[9] S. M. B. P. Samarakoon, M. A. V. J. Muthugala, A. G. B. P. Jayasekara, and M. R. Elara, "Adapting approaching proxemics of a service robot based on physical user behavior and user feedback," *User Model User-adapt Interact*, 2022, doi: 10.1007/s11257-022-09329-8.

[10] F. Haarslev, W. K. Juel, A. Kollakidou, N. Krüger, and L. Bodenhagen, "Context-aware social robot navigation," in *Proceedings of the 18th International Conference on Informatics in Control, Automation and Robotics, ICINCO 2021*, SciTePress, 2021, pp. 426–433. doi: 10.5220/0010554204260433.

[11] P. Aliasghari, A. Taheri, A. Meghdari, and E. Maghsoodi, "Implementing a gaze control system on a social robot in multi-person interactions," *SN Appl Sci*, vol. 2, no. 6, Jun. 2020, doi: 10.1007/s42452-020-2911-0.

[12] T. Kirks, J. Jost, J. Finke, and S. Hoose, "Modelling proxemics for human-technology-interaction in decentralized social-robot-systems," in *Advances in Intelligent Systems and Computing*, 2020. doi: 10.1007/978-3-030-39512-4_24.

[13] E. Repiso, A. Garrell, and A. Sanfeliu, "Adaptive Social Planner to Accompany People in Real-Life Dynamic Environments," *Int J Soc Robot*, 2022, doi: 10.1007/s12369-022-00937-3.

[14] B. Bilen, H. Kivrak, P. Uluer, and H. Kose, "Social Robot Navigation with Adaptive Proxemics Based on Emotions," Jan. 2024, [Online]. Available: http://arxiv.org/abs/2401.17663

[15] J. Karwowski, W. Szynkiewicz, and E. Niewiadomska-Szynkiewicz, "Bridging Requirements, Planning, and Evaluation: A Review of Social Robot Navigation," *Sensors*, vol. 24, no. 9, May 2024, doi: 10.3390/s24092794.

[16] J. Ginés Clavero, F. Martín Rico, F. J. Rodríguez-Lera, J. M. Guerrero Hernandéz, and V. Matellán Olivera, "Impact of decision-making system in social navigation," *Multimed Tools Appl*, vol. 81, no. 3, pp. 3459–3481, Jan. 2022, doi: 10.1007/s11042-021-11454-2.

[17] J. G. Clavero, F. M. Rico, F. J. Rodriguez-Lera, J. M. G. Hernandez, and V. M. Olivera, "Defining Adaptive Proxemic Zones for Activity-aware Navigation," Sep. 2020, [Online]. Available: http://arxiv.org/abs/2009.04770

[18] G. Cohen-Lazry, A. Degani, T. Oron-Gilad, and P. A. Hancock, "Discomfort: an assessment and a model," *Theor Issues Ergon Sci*, vol. 24, no. 4, pp. 480–503, 2023, doi: 10.1080/1463922X.2022.2103201.

[19] K. Marmor *et al.*, "Keeping social distance in a classroom while interacting via a telepresence robot: a pilot study," *Front Neurorobot*, vol. 18, 2024, doi: 10.3389/fnbot.2024.1339000.

[20] S. Kang, S. Yang, D. Kwak, Y. Jargalbaatar, and D. Kim, "Social Type-Aware Navigation Framework for Mobile Robots in Human-Shared Environments," *Sensors*, vol. 24, no. 15, Aug. 2024, doi: 10.3390/s24154862.

[21] Y. S. Kim *et al.*, "Understanding human-robot proxemic norms in construction: How do humans navigate around robots?," *Autom Constr*, vol. 164, Aug. 2024, doi: 10.1016/j.autcon.2024.105455.

[22] A. Bellarbi, A. illah Mouaddib, N. Achour, and N. Ouadah, "A new approach for social navigation and interaction using a dynamic proxemia modeling," *Evol Intell*, vol. 15, no. 3, pp. 2207–2233, Sep. 2022, doi: 10.1007/s12065-021-00633-7.

[23] M. M. E. Neggers, R. H. Cuijpers, P. A. M. Ruijten, and W. A. IJsselsteijn, "Determining Shape and Size of Personal Space of a Human when Passed by a Robot," *Int J Soc Robot*, vol. 14, no. 2, pp. 561–572, Mar. 2022, doi: 10.1007/s12369-021-00805-6.

[24] M. Daza, D. Barrios-Aranibar, J. Diaz-Amado, Y. Cardinale, and J. Vilasboas, "An approach of social navigation based on proxemics for crowded environments of humans and robots," *Micromachines (Basel)*, vol. 12, no. 2, Feb. 2021, doi: 10.3390/mi12020193.

[25] P. Patompak, S. Jeong, I. Nilkhamhang, and N. Y. Chong, "Learning Proxemics for Personalized Human–Robot Social Interaction," *Int J Soc Robot*, vol. 12, no. 1, pp. 267–280, Jan. 2020, doi: 10.1007/s12369-019-00560-9.

[26] E. Torta, R. H. Cuijpers, J. F. Juola, and D. Van Der Pol, "Modeling and testing proxemic behavior for humanoid robots," *International Journal of Humanoid Robotics*, vol. 9, no. 4, Dec. 2012, doi: 10.1142/S0219843612500284.

[27] L. A. Hayduk, "The shape of personal space: An experimental investigation," *Canadian Journal of Behavioural Science*, vol. 13, no. 1, 1981, doi: 10.1037/h0081114.

[28] D. Helbing and P. Molnár, "Social force model for pedestrian dynamics," *Phys Rev E*, vol. 51, no. 5, pp. 4282–4286, 1995, doi: 10.1103/PhysRevE.51.4282.

[29] M. Gérin-Lajoie, C. L. Richards, J. Fung, and B. J. McFadyen, "Characteristics of personal space during obstacle circumvention in physical and virtual environments," *Gait Posture*, vol. 27, no. 2, 2008, doi: 10.1016/j.gaitpost.2007.03.015.

[30] P. T. Singamaneni *et al.*, "A survey on socially aware robot navigation: Taxonomy and future challenges," *Int J Rob Res*, Feb. 2024, doi: 10.1177/02783649241230562.

[31] S. M. B. P. Samarakoon, M. A. V. J. Muthugala, and A. G. B. P. Jayasekara, "A Review on Human–Robot Proxemics," Aug. 01, 2022, *MDPI*. doi: 10.3390/electronics11162490.

[32] M. Argyle and J. Dean, "Eye-contact, distance and affiliation.," *Soins*, vol. 28, 1965, doi: 10.2307/2786027.

[33] T. M. Ciolek and A. Kendon, "Environment and the Spatial Arrangement of Conversational Encounters," *Sociol Inq*, vol. 50, no. 3–4, pp. 237–271, 1980, doi: 10.1111/j.1475-682X.1980.tb00022.x.

[34] M. L. Walters *et al.*, "Close encounters: Spatial distances between people and a robot of mechanistic appearance," in *Proceedings of 2005 5th IEEE-RAS International Conference on Humanoid Robots*, 2005, pp. 450–455. doi: 10.1109/ICHR.2005.1573608.

[35] IEEE Robotics and Automation Society, Robotics Society of Japan, Han'guk Robot Hakhoe, and Institute of Electrical and Electronics Engineers., *Stop! That is Close Enough. How Body Postures Influence Human-Robot Proximity*. 2016.

[36] G. Eresha, M. Häring, B. Endrass, E. André, and M. Obaid, "Investigating the Influence of Culture on Proxemic Behaviors for Humanoid Robots," 2013. [Online]. Available: http://www.aldebaran-robotics.com/en/

[37] C. Mavrogiannis *et al.*, "Core Challenges of Social Robot Navigation: A Survey," *ACM Trans Hum Robot Interact*, Sep. 2023, doi: 10.1145/3583741.

[38] R. Kirby, "Social Robot Navigation," 2010.

[39] H. Hecht, R. Welsch, J. Viehoff, and M. R. Longo, "The shape of personal space," *Acta Psychol (Amst)*, vol. 193, pp. 113–122, Feb. 2019, doi: 10.1016/j.actpsy.2018.12.009.

[40] J. Rios-Martinez, A. Spalanzani, and C. Laugier, "From Proxemics Theory to Socially-Aware Navigation: A Survey," *Int J Soc Robot*, vol. 7, no. 2, pp. 137–153, Apr. 2015, doi: 10.1007/s12369-014-0251-1.

[41] J. N. Bailenson, J. Blascovich, A. C. Beall, and J. M. Loomis, "Equilibrium theory revisited: Mutual gaze and personal space in virtual environments," *Presence: Teleoperators and Virtual Environments*, vol. 10, no. 6, 2001, doi: 10.1162/105474601753272844.

[42] R. C. Newman Ii and D. Pollack, "Proxemics in deviant adolescents," 1973.

[43] K. Klüber and L. Onnasch, "Keep your Distance! Assessing Proxemics to Virtual Robots by Caregivers," Association for Computing Machinery (ACM), Mar. 2023, pp. 193–197. doi: 10.1145/3568294.3580070.

[44] M. M. E. Neggers, S. Belgers, R. H. Cuijpers, P. A. M. Ruijten, and W. A. IJsselsteijn, "Comfortable Crossing Strategies for Robots," *Int J Soc Robot*, May 2024, doi: 10.1007/s12369-024-01127-z.

[45] X. Xu, L. Liying, M. Khamis, G. Zhao, and R. Bretin, "Understanding Dynamic Human-Robot Proxemics in the Case of Four-Legged Canine-Inspired Robots," Feb. 2023, [Online]. Available: http://arxiv.org/abs/2302.10729

[46] H. Yamazoe *et al.*, "Analysis of impressions of robot by changing its motion and trajectory parameters for designing parameterized behaviors of home-service robots," *Intell Serv Robot*, vol. 16, no. 1, pp. 3–18, Mar. 2023, doi: 10.1007/s11370-022-00447-1.

[47] D. Gallo *et al.*, "Investigating the Integration of Human-Like and Machine-Like Robot Behaviors in a Shared Elevator Scenario," Association for Computing Machinery (ACM), Mar. 2023, pp. 192–201. doi: 10.1145/3568162.3576974.

[48] J. Leoste *et al.*, "Keeping distance with a telepresence robot: A pilot study," *Front Educ (Lausanne)*, vol. 7, Jan. 2023, doi: 10.3389/feduc.2022.1046461.

[49] M. Moujahid, D. A. Robb, C. Dondrup, and H. Hastie, "Come Closer: The Effects of Robot Personality on Human Proxemics Behaviours," Sep. 2023, [Online]. Available: http://arxiv.org/abs/2309.02979

[50] K. He, W. P. Chan, A. Cosgun, A. Joy, and E. A. Croft, "Robot Gaze During Autonomous Navigation and Its Effect on Social Presence," *Int J Soc Robot*, 2023, doi: 10.1007/s12369-023-01023-y.

[51] M. M. E. Neggers, R. H. Cuijpers, P. A. M. Ruijten, and W. A. IJsselsteijn, "The effect of robot speed on comfortable passing distances," *Front Robot AI*, vol. 9, Jul. 2022, doi: 10.3389/frobt.2022.915972.

[52] N. E. Neef, S. Zabel, M. Lauckner, and N. S. Otto, "What is Appropriate? On the Assessment of Human-Robot Proxemics for Casual Encounters in Closed Environments," 2022, doi: 10.21203/rs.3.rs-2329385/v1.

[53]     C. Cathcart, M. Santos, S. Park, and N. E. Leonard, "Proactive Opinion-Driven Robot Navigation around Human Movers," Oct. 2022, [Online]. Available: http://arxiv.org/abs/2210.01642

[54]     P. A. M. Ruijten and R. H. Cuijpers, "Do not let the robot get too close: Investigating the shape and size of shared interaction space for two people in a conversation," *Information (Switzerland)*, vol. 11, no. 3, Mar. 2020, doi: 10.3390/info11030147.

[55]     F. W. Siebert, J. Klein, M. Rötting, and E. Roesler, "The Influence of Distance and Lateral Offset of Follow Me Robots on User Perception," *Front Robot AI*, vol. 7, Jun. 2020, doi: 10.3389/frobt.2020.00074.

[56]     L. Miller, J. Kraus, F. Babel, M. Messner, and M. Baumann, "Come Closer: Experimental Investigation of Robots' Appearance on Proximity, Affect and Trust in a Domestic Environment," *Proceedings of the Human Factors and Ergonomics Society Annual Meeting*, vol. 64, no. 1, pp. 395–399, Dec. 2020, doi: 10.1177/1071181320641089.

[57]     H. Lehmann, A. Rojik, and M. Hoffmann, "Should a small robot have a small personal space? Investigating personal spatial zones and proxemic behavior in human-robot interaction," Sep. 2020, [Online]. Available: http://arxiv.org/abs/2009.01818

[58]     D. Sakamoto *et al.*, *Exploring Machine-like Behaviors for Socially Acceptable Robot Navigation in Elevators*. 2022.

[59]     L. Onnasch and E. Roesler, "A Taxonomy to Structure and Analyze Human–Robot Interaction," *Int J Soc Robot*, vol. 13, no. 4, pp. 833–849, Jul. 2021, doi: 10.1007/s12369-020-00666-5.

[60]     M. M. E. Neggers, R. H. Cuijpers, and P. A. M. Ruijten, "Comfortable passing distances for robots," in *Lecture Notes in Computer Science (including subseries Lecture Notes in Artificial Intelligence and Lecture Notes in Bioinformatics)*, Springer Verlag, 2018, pp. 431–440. doi: 10.1007/978-3-030-05204-1_42.

[61]     Björn Petrak; Julia G. Stapels; Katharina Weitz; Friederike Eyssel; Elisabeth André, "To move or not to move Social acceptability of robot proxemics behavior depending on user emotion," 2021.

[62]     K. Charalampous, I. Kostavelis, and A. Gasteratos, "Recent trends in social aware robot navigation: A survey," Jul. 01, 2017, *Elsevier B.V.* doi: 10.1016/j.robot.2017.03.002.